\documentclass[10pt,twocolumn,letterpaper]{article}

\usepackage{cvpr}              

\usepackage[dvipsnames]{xcolor}

\definecolor{cvprblue}{rgb}{0.21,0.49,0.74}
\usepackage[pagebackref,breaklinks,colorlinks,citecolor=cvprblue]{hyperref}
\usepackage{microtype}
\usepackage{graphicx}
\usepackage{subcaption}
\usepackage{caption}
\usepackage{booktabs} 
\usepackage{algorithm}
\usepackage{algpseudocode}
\usepackage{subcaption}
\usepackage{amsmath}
\usepackage{amssymb}
\usepackage{mathtools}
\usepackage{amsthm}
\def\paperID{6} 
\def\confName{CVPR}
\def\confYear{2025}

\title{WQLCP: Weighted Adaptive Conformal Prediction for Robust Uncertainty Quantification Under Distribution Shifts}

\author{Shadi Alijani \hspace{1em} Homayoun Najjaran \\
University of Victoria \hspace{2em}\\
{\tt\small\{shadialijani,najjaran\}@uvic.ca}
}

\begin{document}
\maketitle
\begin{abstract}
Conformal prediction (CP) provides a framework for constructing prediction sets with guaranteed coverage, assuming exchangeable data. However, real-world scenarios often involve distribution shifts that violate exchangeability, leading to unreliable coverage and inflated prediction sets. To address this challenge, we first introduce Reconstruction Loss-Scaled Conformal Prediction (RLSCP), which utilizes reconstruction losses derived from a Variational Autoencoder (VAE) as an uncertainty metric to scale score functions. While RLSCP demonstrates performance improvements, mainly resulting in better coverage, it quantifies quantiles based on a fixed calibration dataset without considering the discrepancies between test and train datasets in an unexchangeable setting. In the next step, we propose Weighted Quantile Loss-scaled Conformal Prediction (WQLCP), which refines RLSCP by incorporating a weighted notion of exchangeability, adjusting the calibration quantile threshold based on weights with respect to the ratio of calibration and test loss values. This approach improves the CP-generated prediction set outputs in the presence of distribution shifts. Experiments on large-scale datasets, including ImageNet variants, demonstrate that WQLCP outperforms existing baselines by consistently maintaining coverage while reducing prediction set sizes, providing a robust solution for CP under distribution shifts.
\end{abstract}    
\section{Introduction}
\label{sec:intro}
Deep learning models have demonstrated remarkable performance across a wide range of tasks, achieving state-of-the-art results in domains such as image recognition , scene understanding, and automated decision systems \cite{silver2017mastering, brown2020language, dosovitskiy2020image}. However, in safety-critical applications—from autonomous navigation to industrial inspection—high accuracy alone is insufficient; understanding and quantifying uncertainty is equally crucial to ensure reliable and trustworthy predictions \cite{amodei2016concrete, ovadia2019can, fayyad2024empirical}. In practical implementations of uncertainty quantification (UQ), particularly when deploying vision systems to guide high-stakes decisions, the ability to interpret model predictions and rigorously quantify uncertainty can significantly enhance both algorithmic robustness and real-world applicability. For instance, in tasks like classification or image segmentation, confidence estimates directly influence downstream actions, operational safety, and error mitigation strategies. This underscores the need for robust, interpretable, and calibrated uncertainty measures \cite{begoli2019need, nair2020exploring, zhou2022high, liu2023bayesian, wang2023uncertainty}, ensuring that vision-based models align with human expectations of reliability while operating in dynamic, open-world environments.

CP methods \cite{vovk2005algorithmic} offer a promising solution by providing model-generated uncertainty estimates with finite-sample statistical guarantees. These methods generate prediction sets that, with a user-defined confidence level, are statistically guaranteed to contain the true label, making them suitable for high-stakes decision-making scenarios \cite{sadinle2019least, romano2020classification, angelopoulos2023conformal}. Despite their proven success, the performance of CP methods has predominantly been demonstrated in well-curated, in-distribution settings. However, real-world data streams are often affected by various types of distribution shifts, violating the exchangeability assumption underlying CP. These shifts can arise due to discrepancies between training and test distributions, such as domain changes or sampling biases, which pose challenges for maintaining reliable coverage and prediction set sizes. In such cases, conventional exchangeability-based CP methods often fail to adapt effectively to these changes, leading to degraded performance \cite{bhatnagar2023improved, kasa2023empirically}. Consequently, addressing distribution shifts within the CP framework has become a central focus of recent research efforts.

Although many methods have been proposed, addressing the core challenges in CP remains an ongoing effort. For example, some CP approaches, such as \cite{tibshirani2019conformal, roth2022uncertain, barber2023conformal, cauchois2024robust}, have shown the ability to handle distribution shifts. However, these methods often rely on simplifying assumptions about the nature of distribution shifts—such as uniformity or limited scope—mainly due to high computational demands. Furthermore, many existing CP methods use simplistic nonconformity metrics that are insufficient for high-dimensional image data \cite{belhasin2023principal}. 

\textbf{Contributions:}
Our method builds on using reconstruction-based uncertainty metrics, addressing gaps identified in prior work. By employing reconstruction losses, we bypass the need for strong assumptions about the nature of distribution shifts. This aligns with recent findings that emphasize the importance of robust and adaptive UQ in CP frameworks \cite{zhou2024conformal}. We propose two novel methods:

\begin{itemize}
    \item \textbf{Reconstruction-Loss-Scaled Conformal Prediction (RLSCP)}: A foundational approach that dynamically scales prediction set sizes using VAE-derived reconstruction losses. RLSCP directly links reconstruction uncertainty to conformal scores, ensuring larger prediction sets for high-uncertainty inputs (e.g., distribution-shifted or ambiguous images) without requiring prior assumptions about shift uniformity. This mitigates coverage degradation in safety-critical vision tasks.
    
    \item \textbf{Weighted Quantile Loss-scaled for Conformal Prediction (WQLCP)}: Refinement of RLSCP that introduces a weighted quantile calibration framework. By approximating the calibration-test reconstruction loss ratio, WQLCP adaptively reweights calibration samples to: \begin{itemize}
        \item Recover target coverage rates under severe distribution shifts
        \item Minimize prediction set sizes by prioritizing high-uncertainty calibration samples
    \end{itemize}
    while preserving the finite-sample statistical guarantees of CP.
\end{itemize}

Together, these methods bridge the gap between heuristic UQ and distributionally robust CP, achieving state-of-the-art performance on complex image datasets with theoretical and empirical rigor.


\section{Background}
CP has demonstrated its versatility across various domains. For example, generic treatments of CP methods are detailed in \cite{lekeufack2024conformal}, while their use in robotics for trajectory optimization \cite{lindemann2023safe} and 3D vision tasks \cite{yang2023object, gao2024closure} underscores their wide applicability. CP has also been explored with large language models (LLMs) in applications like uncertainty-aware decision-making \cite{kumar2023conformal, quach2023conformal} and LLM-powered robotics \cite{ren2023robots}. In decentralized systems, CP has been applied to multi-robot planning \cite{wang2024safe} and improving decision-making in LLM-driven tasks \cite{vishwakarma2024improving}. In the context of image recognition, recent studies have applied CP to calibrate image segmentation models \cite{brunekreef2024kandinsky}, quantify post-hoc uncertainties in semantic segmentation \cite{mossina2024conformal}, and evaluate CP methods under challenging conditions, such as distribution shifts and long-tailed data \cite{kasa2023empirically}. CP has also been utilized to enhance reliability in medical image classification tasks, providing rigorous statistical guarantees through reliable-region-based conformal prediction (RR-CP) methods \cite{zhang2023rr}. Additionally, it has been applied to disease severity grading, such as spinal stenosis diagnosis from lumbar spine MRIs \cite{lu2022improving}, and to improve trust in AI-driven diagnostic tools \cite{vazquez2022conformal, gao2024closure}. Research on foundation models as conformal predictors \cite{fillioux2024foundation} further reinforces the adaptability of CP to modern medical imaging systems, where scalability and robustness are critical. These advances underscore the versatility and potential of CP in addressing the challenges posed by machine-assisted medical decision-making.

\subsection{CP under distribution shifts}
Recent advancements in CP have focused on enhancing its robustness under distribution shifts and expanding its applicability beyond conventional assumptions of data exchangeability. \cite{tibshirani2019conformal} introduced methods for handling covariate shifts through likelihood ratio reweighting, while \cite{cauchois2024robust} extended this approach with robust optimization techniques. \cite{roth2022uncertain} explored multicalibration strategies to manage diverse test distributions, \cite{barber2023conformal} developed generalized reweighting techniques for mild distribution shifts. Additionally, CP has been linked with adversarial online learning in works such as \cite{gibbs2024conformal}, \cite{angelopoulos2023conformal}, and \cite{zhang2024discounted}, which employ regret minimization algorithms for dynamic adaptation under continual distribution shifts.

Despite these advancements, critical challenges remain unresolved. Approaches such as likelihood ratio reweighting for covariate shifts \cite{tibshirani2019conformal} struggle with computational inefficiencies and inaccuracies in high-dimensional settings, limiting their applicability to more complex shifts. Robust optimization methods \cite{cauchois2024robust} and multi-calibration techniques \cite{roth2022uncertain} extend CP to broader scenarios but require prior knowledge or assumptions about the test domain, which are often impractical in real-world applications. Generalized weighting approaches \cite{barber2023conformal} offer greater flexibility but lack principled methods for selecting appropriate weights, particularly in scenarios involving severe or multi-modal distribution shifts. These gaps highlight the need for methods that adapt dynamically to the test distribution without relying on assumptions or computationally expensive processes. Our proposed WQLC addresses these challenges by using test-specific uncertainty metrics to dynamically adjust calibration quantiles, ensuring robust performance under complex and unknown distribution shifts.

\subsection{Fundamentals of CP}
To establish the foundation, we introduce the basic principles of CP \cite{vovk2005algorithmic, shafer2008tutorial, angelopoulos2023conformal, tibshirani2023advanced}. CP provides a framework for constructing prediction sets with a guaranteed coverage probability \(1 - \alpha\), assuming i.i.d. (independent and identically distributed) data. 

Given a dataset \(\mathcal{D} = \{(x_i, y_i)\}_{i=1}^n \subseteq \mathcal{X} \times \mathcal{Y}\), sampled i.i.d. from an unknown distribution \(D\), the objective of CP is to construct a set-valued function \(\mathcal{C}_D : \mathcal{X} \to 2^{\mathcal{Y}}\) such that:
\begin{equation}
P_{(x, y) \sim D}\big(y \in \mathcal{C}_D(x)\big) \geq 1 - \alpha.
\end{equation}

This ensures that, for a fresh sample \((x, y) \sim D\), the prediction set \(\mathcal{C}_D(x)\) covers the true label \(y\) with high probability. However, this guarantee alone is not useful unless the prediction sets are small, as predicting the entire label space \(\mathcal{C}_D(x) = \mathcal{Y}\) trivially satisfies the coverage guarantee. Therefore, CP seeks to balance between satisfying the coverage guarantee and minimizing the size of prediction sets.

To achieve this, CP utilizes a scoring function \(s : \mathcal{X} \times \mathcal{Y} \to \mathbb{R}^+\), typically derived from a trained machine learning model, that assigns a score \(s(x, y')\) to each covariate-label pair \((x, y')\). Higher scores indicate higher confidence that \(y'\) is the correct label for \(x\). The prediction set is then constructed as:
\begin{equation}
\mathcal{C}_D(x; \tau_D) := \{y' \in \mathcal{Y} : s(x, y') \geq \tau_D\}.
\end{equation}

where \(\tau_D\) is a threshold determined using the calibration dataset.

Under the i.i.d. assumption, the threshold \(\tau_D\) is chosen as the $\alpha \left(1 + \frac{1}{n}\right)$-quantile of the empirical scores \(\{s(x_i, y_i)\}_{i=1}^n\), ensuring the coverage objective is satisfied. This process, which splits the data into training and calibration sets, is known as Split Conformal Prediction (SplitCP).

While CP provides marginal coverage guarantees, it does not ensure conditional coverage under distribution shifts or other challenging scenarios. This motivates the development of advanced methods to adapt CP to these settings, which are the focus of this work.


\section{Methods}
We propose two complementary methods, which we elaborate on in detail in the following sections.
 
\subsection{Primary Approach}
\begin{figure}[ht]
\vskip 0.2in
\begin{center}
\centerline{\includegraphics[width=\columnwidth]{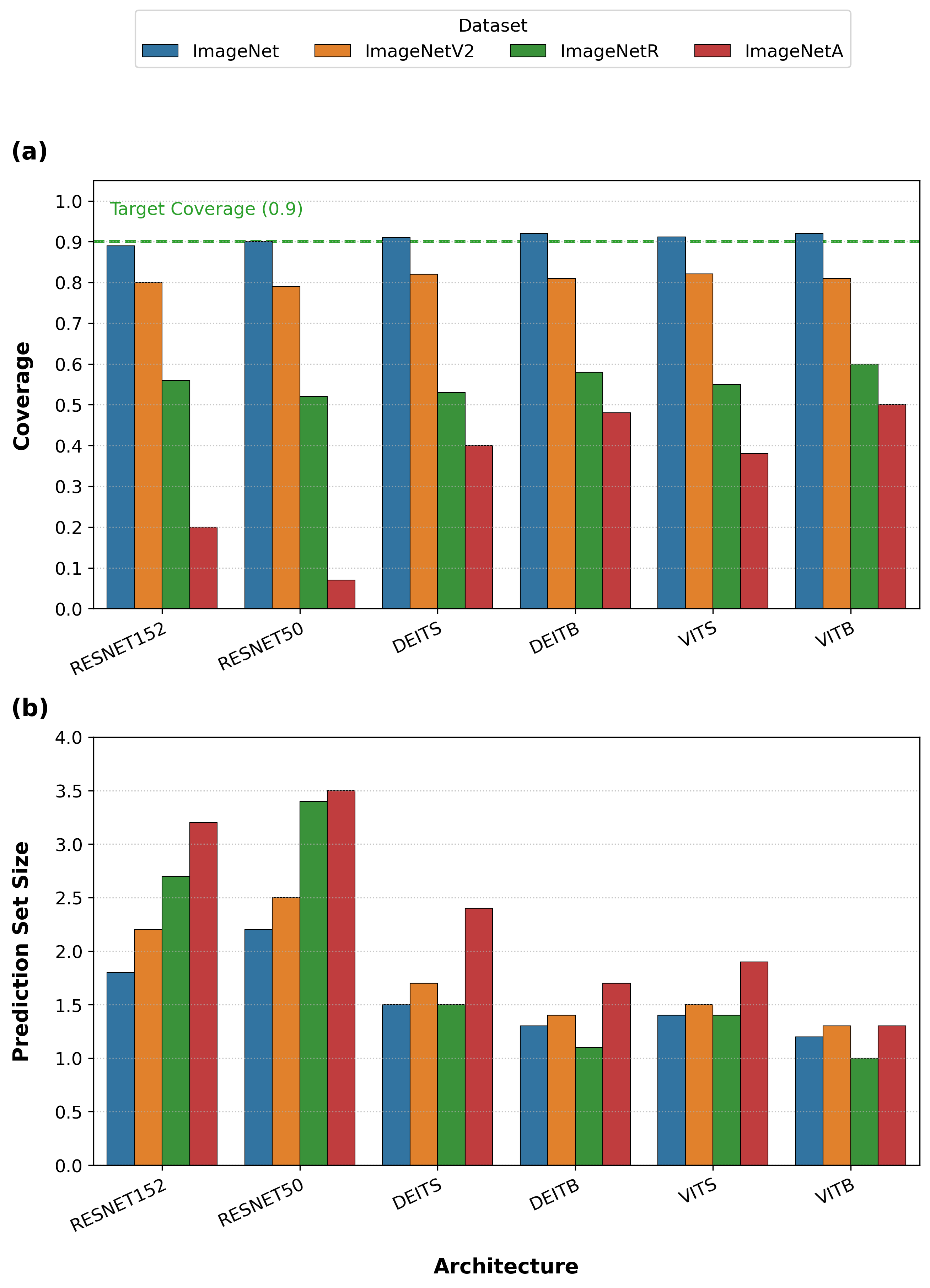}}
\caption{(a) CP coverage violations under distribution shifts. While transformer-based networks (ViTs, DeiTs) perform better compared to CNNs (ResNet50, ResNet152), they still fail to maintain coverage in distribution-shifted scenarios. The desired coverage level is set to 0.90. (b) Prediction set sizes: The diagram illustrates larger prediction set sizes for ImageNet variant datasets.}
\label{CP violation under distribution shifts}
\end{center}
\vskip -0.2in
\end{figure}

Based on our initial experiments, illustrated in Figure \ref{CP violation under distribution shifts}, and supported by recent research \cite{alijani2024vision}, ViT-based backbones demonstrate better performance in handling distribution shifts compared to Convolutional Neural Networks (CNNs). However, even ViT-based models with post-hoc CP methods struggle to maintain coverage and exhibit prediction set size violations under such conditions. In addition to empirical evaluations, building upon insights from recent studies, our core idea is to apply uncertainty-based scaling to the score function to enhance CP’s performance in these scenarios. Unlike most previous methods that focus on estimating the base model’s uncertainty at individual covariates $x$, our approach evaluates uncertainty across the entire of the distribution-shifted test datasets. This strategy aggregates the localized uncertainties at each $x$, by considering uncertainty across the entire test dataset. Instead of prioritizing local adaptivity with respect to $x$, as in previous research for mainly time series data, our method with using ViTs as backbone models, focuses on adaptivity to the broader distribution shift, capturing global evaluation to have a more effective handling of the challenges presented by unseen test data.

\subsubsection{Rescaling Conformal Scores Using VAE-Derived Uncertainty}
A key design choice in CP frameworks is the uncertainty metric. Following \cite{kingma2013auto}, we adopt the reconstruction loss of a variational auto encoder (VAE) as our uncertainty metric. This choice is motivated by the VAE's ability to quantify uncertainty through its probabilistic framework, which maximizes the Evidence Lower Bound (ELBO) to approximate latent structure in data distributions. The ELBO balances two competing objectives:
\begin{align}
    \text{ELBO} &= \underbrace{\mathbb{E}_{q(z|x)}[\log p(x|z)]}_{\text{Reconstruction}} - \underbrace{\text{KL}(q(z|x) \parallel p(z))}_{\text{Regularization}}. 
    \label{eq:elbo}
\end{align}
where the first term, \(-\mathbb{E}_{q(z|x)}[\log p(x|z)]\) (reconstruction loss), measures the model's ability to reconstruct the input \(x\) from its latent representation \(z\). The second term, the Kullback-Leibler (KL) divergence, regularizes the latent space by penalizing deviations of the approximate posterior \(q(z|x)\) from the prior \(p(z)\).

\begin{figure}[ht]
\vskip 0.2in
\begin{center}
\centerline{\includegraphics[width=\columnwidth]{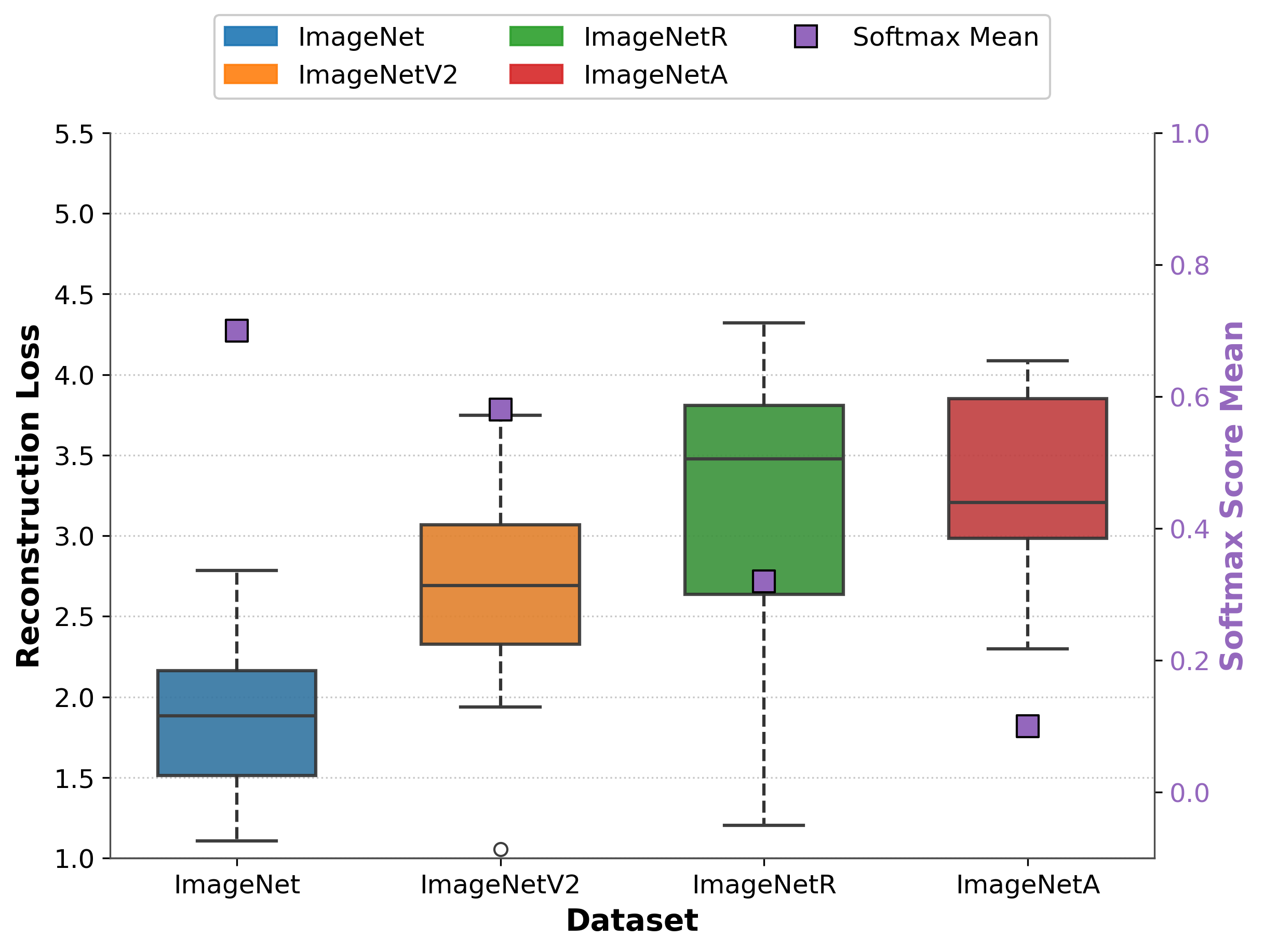}}
\caption{Comparison of reconstruction loss and softmax mean of the true label across ImageNet, ImageNetV2, ImageNetR, and ImageNetA.}
\label{Box Plot Of Reconstruction Loss With Softmax Scores Overlay}
\end{center}
\vskip -0.2in
\end{figure}

While the KL divergence ensures latent space regularity \citep{asperti2020balancing}, the reconstruction loss directly reflects the model's confidence in reproducing the input data. Higher reconstruction loss implies greater difficulty in modeling \(x\), which we interpret as higher epistemic uncertainty (i.e., uncertainty due to insufficient knowledge about the data distribution). This makes \(\mathcal{L}_{\text{rec}}(x)\) a natural candidate for CP, where quantifying input-specific uncertainty is critical for constructing adaptive prediction intervals. Figure~\ref{Box Plot Of Reconstruction Loss With Softmax Scores Overlay} shows the reconstruction loss values for  samples averaged over each dataset. Our empirical experiment demonstrates that $\mathcal{L}_{\text{rec}}(x)$ correlates with the severity of the distribution shift in the ImageNet variant. In other words, larger shifts with low softmax scores are strongly correlated with higher reconstruction losses, indicating higher uncertainty in the base model.

Our method leverages this relationship by aggregating reconstruction losses across the entire test dataset to provide a global perspective of uncertainty. This aggregation enables the detection of distribution shifts comprehensively, rather than relying only on localized uncertainty based on individual covariates $x$.

\begin{figure*}[t]
\vskip 0.2in
\begin{center}
\centerline{\resizebox{1.0\textwidth}{!}{\includegraphics{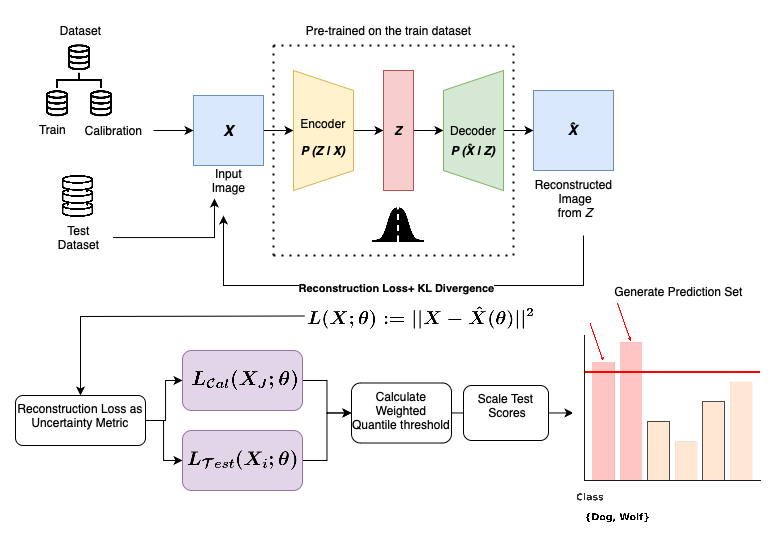}}}
\caption{Overview of the WQLCP framework. The model leverages VAEs to compute reconstruction losses for uncertainty estimation, applying weighted quantile and scaling test scores to refine conformal prediction.}
\label{WQLCP}
\end{center}
\vskip -0.2in
\end{figure*}

\subsubsection{RLSCP: Mathematical Principles}
Having established the reconstruction loss distribution, we now employ it as an uncertainty metric to adaptively scale CP scores. Let $\mathcal{L}_{\text{test}} = \{\mathcal{L}(x_1), \dots, \mathcal{L}(x_N)\}$ denote the reconstruction losses computed over the test dataset, and let $q_k(\cdot)$ represent the $k$-th quantile operator. We define the \emph{reconstruction loss quantile threshold} as:

\begin{equation}
\label{eq:rl_quantile}
RL_{\text{test}} = q_{1-\alpha}(\mathcal{L}_{\text{test}})
\end{equation}

where $\alpha \in (0,1)$ corresponds to the user-specified error tolerance level in CP. This threshold enables adaptive scale of prediction sets through the following scaled score criterion:

\begin{equation}
\label{eq:scaled_score}
C_{\text{test}}(x; \tau_{\text{test}}) = \left\{y' \in \mathcal{Y} : s(x, y') \cdot \max\left(1, RL_{\text{test}}\right) \geq \tau_{\text{test}}\right\}
\end{equation}

Here, $\tau_{\text{test}}$ is the CP threshold. The $\max(1, \cdot)$ operator ensures two critical properties:

\begin{itemize}
    \item \textbf{Compatibility}: Preserves the original SplitCP prediction sets when $RL_{\text{test}} \leq 1$
    \item \textbf{Adaptivity}: Scales the score function proportionally to reconstruction loss when $RL_{\text{test}} > 1$
\end{itemize}

Under distribution shift, test samples typically exhibit elevated reconstruction losses due to decoder miscalibration. The $RL_{\text{test}}$ quantile acts as a shift severity estimator -- larger shifts produce higher scaling factors. This scaling relaxes the effective threshold for label inclusion, systematically enlarging prediction sets to counter SplitCP's tendency toward under-coverage in shifted distributions. The adaptive trade-off between set size and UQ thereby maintains nominal coverage guarantees despite covariate shift.
\subsection{Weighted Quantile Optimization for Improved Uncertainty Calibration}
While the proposed RLSCP method improves coverage guarantees for CP under distribution shifts by scaling based on the $RL_{\text{test}}$, it increases prediction set sizes, a trade-off that can reduce practical usability in real-world applications. Although RLSCP demonstrates measurable improvements on shifted benchmarks like ImageNet-V2/R/A, its dependence on both fixed calibration processes and static base models motivates the need for test-distribution-aware calibration.

\subsubsection{WQLCP Algorithm}
\label{subsubsec:wqlcp_framework}

We introduce WQLCP, which integrates test-specific uncertainties into the calibration process through reconstruction loss analysis. For a VAE with parameters $\theta$, we define the reconstruction loss as:
\begin{equation}
    \mathcal{L}(x; \theta) := \|x - \hat{x}(\theta)\|^2,
    \label{eq:recon_loss}
\end{equation}
where $\hat{x}(\theta)$ represents the reconstructed input. Our key insight establishes reconstruction loss ratios as distribution alignment: lower values of $\mathcal{L}(x; \theta)$ suggest strong calibration-test alignment (in-distribution samples), while higher values signal distribution shifts (out-of-distribution samples). This relationship enables the definition of calibration weights proportional to the ratio of calibration-to-test reconstruction losses:
\begin{equation}
    w(x_j) \propto \frac{\mathcal{L}_{\text{cal}}(x_j)}{\mathcal{L}_{\text{test}}(x_i) + \epsilon},
    \label{eq:weight_ratio}
\end{equation}
where $\mathcal{L}_{\text{cal}}$ and $\mathcal{L}_{\text{test}}$ denote reconstruction losses for calibration ($x_j$) and test ($x_i$) samples respectively, with $\epsilon$ ensuring numerical stability.

\subsubsection{Algorithmic Implementation}
\label{subsubsec:algorithm}

As visualized in Figure~\ref{WQLCP}, WQLCP inspired by established weighted CP methods~\cite{tibshirani2019conformal,barber2023conformal} by incorporating ViT-VAE reconstruction loss ratios. Building on SplitCP~\cite{sadinle2019least}, the framework employs nonconformity scores that quantify test-calibration divergence, adaptively weighting calibration samples based on their dissimilarity to the test distribution. The weighted quantile estimation follows:

\begin{equation}
    \hat{q} = \inf\left\{q : \sum_{j=1}^n w(x_j)\mathbb{I}\{s_j \leq q\} \geq (1-\alpha)\sum_{j=1}^n w(x_j)\right\}
    \label{eq:weighted_quantile}
\end{equation}
where $s_j$ denotes conformal scores. This dual mechanism combines RLSCP's coverage-preserving score scaling with uncertainty-aware quantile adjustment, achieving simultaneous maintenance of coverage guarantees and reduction of prediction set sizes. The complete implementation details are provided in Algorithm~\ref{alg:wqlcp}.

\begin{algorithm}[H]
\small 
\caption{Weighted Quantile Loss-Scaled Conformal Prediction (WQLCP)}
\label{alg:wqlcp}
\begin{algorithmic} 
\Require 
    Test dataset $\mathcal{D}_{\text{test}} = \{x_i\}_{i=1}^n$, Calibration dataset $\mathcal{D}_{\text{cal}} = \{x_j\}_{j=1}^m$, Confidence level $1 - \alpha$.
\Ensure Prediction sets $\{\mathcal{C}(x_i)\}_{i=1}^n$.

\State \textbf{Step 1: Compute Reconstruction Losses} 
\For{$x_j \in \mathcal{D}_{\text{cal}}$} Compute $L_{\text{cal}}(x_j; \theta)$ using VAE. \EndFor
\For{$x_i \in \mathcal{D}_{\text{test}}$} Compute $L_{\text{test}}(x_i; \theta)$ using VAE. \EndFor

\State \textbf{Step 2: Calculate Weights} 
\For{$x_j \in \mathcal{D}_{\text{cal}}$} Compute $w(x_j) \propto \frac{L_{\text{cal}}(x_j; \theta)}{L_{\text{test}}(x_i; \theta) + \epsilon}$. \EndFor

\State \textbf{Step 3: Compute Weighted Quantile Threshold}
\\
$\hat{q} = \inf\left\{q : \sum_{j=1}^n w(x_j)\mathbb{I}\{s_j \leq q\} \geq (1-\alpha)\sum_{j=1}^n w(x_j)\right\}$

\State \textbf{Step 4: Scale Test Scores} 
\For{$x_i \in \mathcal{D}_{\text{test}}$} Compute $s_{\text{scaled}}(x_i, y) = s(x_i, y) \cdot \max(1,RL_{\text{test}})$. \EndFor

\State \textbf{Step 5: Generate Prediction Sets} 
\For{$x_i \in \mathcal{D}_{\text{test}}$} Construct $\mathcal{C}(x_i) = \{y : s_{\text{scaled}}(x_i, y) \geq \tau_{\text{test}}\}$. \EndFor

\State \textbf{Step 6: Return Prediction Sets} Return $\{\mathcal{C}(x_i)\}$.
\end{algorithmic}
\end{algorithm}

\section{Experimental Setups and Results}

To ensure a comprehensive evaluation, we conduct baseline experiments across diverse backbone architectures, incorporating both convolutional and transformer-based networks.

For CNN architectures, we utilize ResNet-50 and ResNet-152, pretrained on ImageNet-1k \cite{deng2009imagenet}. For transformer-based models, we employ ViT-S/B, which use 16$\times$16 patch embeddings and are pretrained on ImageNet-21k \cite{dosovitskiy2020image}. Additionally, we evaluate DeiT-S/B, which incorporate knowledge distillation during ImageNet-1k training \cite{touvron2022deit}. Our proposed methods leverage VAEs with ResNet and ViT backbones, integrating reconstruction loss-based uncertainty estimation into CP.

All models are trained for 100 epochs using the AdamW optimizer with a learning rate of $10^{-4}$ and a batch size of 256. For distribution-shifted ImageNet benchmarks, we partition the original ImageNet development set into two distinct subsets: 25,000 calibration samples and a separate validation set.

\subsection{Evaluation Metrics}
We assess model performance using three primary metrics:

\textbf{Coverage:} Measures the proportion of test samples where the true label is included in the prediction set, defined as:
\begin{equation}
    \text{Coverage} = \frac{1}{N}\sum_{i=1}^N \mathbb{I}(y_i \in \mathcal{C}(x_i))
    \label{eq:coverage}
\end{equation}

\textbf{Prediction Set Size:} Represents the average number of labels in the prediction set:
\begin{equation}
    \text{Set Size} = \frac{1}{N}\sum_{i=1}^N |\mathcal{C}(x_i)|
    \label{eq:set_size}
\end{equation}

\textbf{Shift Severity:} Quantifies distribution shifts using the normalized reconstruction loss, computed as:
\begin{equation}
    \text{Shift Severity} = \frac{\mathcal{L}_{\text{rec}}(x)}{\mathbb{E}[\mathcal{L}_{\text{rec}}]_{\text{IN}}}
    \label{eq:shift_severity}
\end{equation}
where \( \mathcal{L}_{\text{rec}}(x) \) denotes the sample's reconstruction error, and \( \mathbb{E}[\mathcal{L}_{\text{rec}}]_{\text{IN}} \) represents the mean reconstruction error over in-distribution (IN) data.

\subsection{Datasets}
\label{subsec:setup}
We evaluate our methods on three benchmark distribution shifts derived from ImageNet dataset \cite{deng2009imagenet}:
\begin{itemize}
    \item \textbf{ImageNetV2} \cite{recht2019imagenet}: Contains 10,000 images collected with the original class distribution but under natural distribution shifts (lighting, pose variations). Serves as a mild shift benchmark.
    
    \item \textbf{ImageNetR} \cite{hendrycks2021many}: 30,000 artistic renditions (paintings, sketches) of 200 ImageNet classes. Tests robustness to domain gaps in visual style.
    
    \item \textbf{ImageNetA} \cite{hendrycks2021natural}: 7,500 adversarially filtered natural images that cause model misclassifications. Represents extreme distribution shifts.
\end{itemize}


\begin{table*}[ht]
\centering
\caption{Different CP methods performance for ImageNet, and ImageNet distribution shifted datasets. The results are presented as Average coverage / Average setsize}
\label{tab:baseline_comparison}
\resizebox{\textwidth}{!}{%
\begin{tabular}{@{}lcccccc@{}}
\toprule
\textbf{Method}           & \textbf{Backbone} & \textbf{ImageNet} & \textbf{ImageNetV2} & \textbf{ImageNetR} & \textbf{ImageNetA} \\ \midrule
\multicolumn{6}{c}{\textbf{Baselines}} \\ \midrule
Naive           & RN50             & 0.7523 / 4.5123      & 0.6462 / 4.4837        & 0.3521 / 5.3459       & 0.0152 / 5.5342        \\
                & RN152            & 0.7525 / 4.2321      & 0.6546 / 4.1487        & 0.3869 / 4.8132       & 0.1211 / 5.1028        \\
                & ViT-S            & 0.7721 / 3.8528      & 0.6689 / 3.7342        & 0.4034 / 4.3124       & 0.3568 / 4.7341        \\
                & ViT-B            & 0.7723 / 3.4457      & 0.6732 / 3.3759        & 0.4521 / 3.9123       & 0.3818 / 4.1128        \\
                & DeiT-S           & 0.7625 / 3.2124      & 0.6834 / 3.1187        & 0.3741 / 3.6348       & 0.3021 / 3.9234        \\
                & DeiT-B           & 0.7713 / 3.0021      & 0.6924 / 3.0018        & 0.4041 / 3.3234       & 0.3123 / 3.7014        \\
\midrule
THR \cite{sadinle2019least}           & RN50             & 0.8921 / 2.0523      & 0.8024 / 2.4618        & 0.5023 / 3.3728       & 0.0291 / 3.0712        \\
                & RN152            & 0.9021 / 1.7921      & 0.8123 / 2.0828        & 0.5293 / 2.7418       & 0.1723 / 3.0823        \\
                & ViT-S            & 0.9028 / 1.3728      & 0.8012 / 1.4821        & 0.5188 / 1.3223       & 0.3723 / 1.7521        \\
                & ViT-B            & 0.8998 / 1.1928      & 0.8028 / 1.2221        & 0.5793 / 0.9523       & 0.4721 / 1.2023        \\
                & DeiT-S           & 0.9023 / 1.3421      & 0.8131 / 1.6123        & 0.5083 / 1.4028       & 0.3928 / 2.2490        \\
                & DeiT-B           & 0.8127 / 1.2828      & 0.8121 / 1.3221        & 0.5498 / 1.0123       & 0.4623 / 1.4123        \\
\midrule
APS \cite{romano2020classification}            & RN50             & 0.8993 / 9.0623      & 0.8623 / 15.8021       & 0.7123 / 26.5523      & 0.2023 / 16.1321       \\
                & RN152            & 0.8968 / 6.3723      & 0.8621 / 11.6021       & 0.7223 / 22.7223      & 0.4023 / 15.8623       \\
                & ViT-S            & 0.9014 / 4.5321      & 0.8723 / 8.1723        & 0.7723 / 26.3021      & 0.5723 / 13.9521       \\
                & ViT-B            & 0.8923 / 4.6421      & 0.8789 / 9.1021        & 0.7970 / 27.4721      & 0.7023 / 16.3921       \\
                & DeiT-S           & 0.9039 / 90.531     & 0.9023 / 132.91      & 0.8013 / 73.5323      & 0.7190 / 69.3821       \\
                & DeiT-B           & 0.8987 / 11.602     & 0.8721 / 18.162       & 0.7598 / 43.1223      & 0.6309 / 30.6623       \\
\midrule
RAPS \cite{angelopoulos2020uncertainty}           & RN50             & 0.9002 / 3.7821      & 0.7671 / 2.2521        & 0.4823 / 3.4923       & 0.0221 / 3.0821        \\
                & RN152            & 0.9021 / 2.9821      & 0.7823 / 2.0123        & 0.5223 / 3.1421       & 0.1723 / 3.0523        \\
                & ViT-S            & 0.9023 / 1.7223      & 0.8323 / 2.0923        & 0.5923 / 3.5921       & 0.4723 / 3.2923        \\
                & ViT-B            & 0.9021 / 1.5423      & 0.8423 / 1.8923        & 0.6721 / 3.2023       & 0.5321 / 3.1321        \\
                & DeiT-S           & 0.8999 / 2.0921      & 0.8421 / 2.6421        & 0.5923 / 5.2521       & 0.5099 / 5.1923        \\
                & DeiT-B           & 0.8991 / 1.5921      & 0.8323 / 1.8923        & 0.6223 / 3.6623       & 0.4809 / 3.2723        \\
\midrule
\multicolumn{6}{c}{\textbf{Proposed Methods}} \\
\midrule
RLSCP (Ours)    & RN-VAE           & 0.9201 / 2.5001      & 0.9102 / 7.8003        & 0.7503 / 21.002       & 0.6002 / 50.512        \\
                & ViT-VAE          & 0.9301 / 2.2003      & 0.9304 / 7.0002        & 0.7992 / 16.703       & 0.7703 / 28.514        \\
\midrule
WQLCP (Ours)    & RN-VAE           & 0.9402 / 2.0002      & 0.9292 / 8.9001        & 0.7773 / 11.112       & 0.6501 / 12.122        \\
                & ViT-VAE          & \textbf{0.9503 / 1.8001 }     & \textbf{0.9403 / 6.7003}        & \textbf{0.8502 / 8.302}       & \textbf{0.7402 / 9.501}        \\
\midrule
\multicolumn{6}{c}{\textbf{Comparison Against SOTA}} \\
\midrule                                          
WCP \cite{tibshirani2019conformal} & ---         & 0.8801 / 10.5002      & 0.8702 / 18.8002        & 0.7503 / 8.5001       & 0.6503 / 7.3002        \\
SSCP \cite{seedat2023improving}  & ---           & 0.9001 / 11.2002      & 0.8603 / 17.0003        & 0.7703 / 9.7001       & 0.6702 / 8.6003        \\  
\bottomrule
\end{tabular}%
}
\end{table*}

\begin{figure*}[ht]
    \centering
    \includegraphics[width=1.0\textwidth]{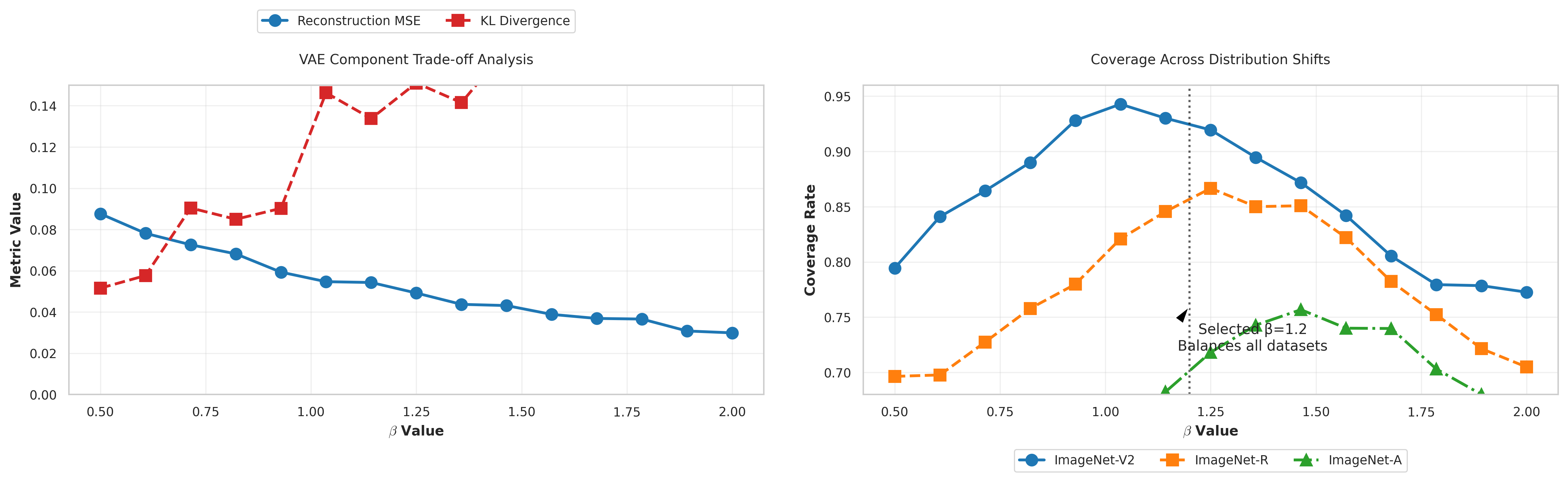}
    \caption{$\beta$-VAE ablation study: (Left) Reconstruction MSE loss vs KL divergence trade-off. (Right) Coverage vs $\beta$ on ImageNetA, showing optimal $\beta=1.2$ achieves 0.7402 coverage.}
    \label{fig:beta_ablation}
\end{figure*}

\subsection{Quantitative Performance Comparison}
As demonstrated in Table~\ref{tab:baseline_comparison}, we observe that across all datasets, WQLCP consistently matches or outperforms SOTA (WCP and SSCP) in coverage while achieving a significantly smaller prediction set size. On mild shifts (ImageNetV2, ImageNetR), WQLCP achieves \textbf{0.9403 / 6.7003} and \textbf{0.8502 / 8.302}, respectively, closely matching or outperforming SSCP (\textbf{0.8603 / 17.0003}, \textbf{0.7703 / 9.7001}) and WCP (\textbf{0.8702 / 18.8002}, \textbf{0.7503 / 8.5001}). The substantial set size reduction while maintaining high coverage highlights WQLCP’s efficiency in adapting to distribution shifts.  

Under extreme adversarial shifts (ImageNetA), WQLCP(RN-VAE) achieves \textbf{0.6501 / 12.122}; despite a slight coverage drop compared to SSCP, WQLCP (ViT-VAE) ensures good coverage while providing reliable prediction sets in challenging scenarios. WQLCP/ViT-VAE strikes a practical balance, attaining \textbf{0.7402} coverage with a set size of \textbf{9.501}, keeping uncertainty manageable while outperforming APS/ViT-B by \textbf{5.4\%} in coverage and reducing set size by \textbf{42.0\%}. This considerable set size reduction is attributed to two key factors: (1) WQLCP updates the underlying neural network on new data, allowing dynamic adaptation to shifts, and (2) unlike SOTA methods that focus on learning a threshold, WQLCP additionally refines conformal scores, leading to more efficient and calibrated prediction sets.
\subsection{Key Technical Insights}
Three principal insights emerge from our technical analysis. First, we validate the strong correlation between VAE reconstruction loss and performance degradation under distribution shift. On the severely shifted ImageNetA benchmark, where reconstruction losses are highest as demonstrated before, RN-VAE suffers a coverage drop of $\Delta = 0.290$ (ImageNet: 94.0\% $\rightarrow$ ImageNetA: 65.0\%), while ViT-VAE exhibits a smaller drop of $\Delta = 0.210$ (95.0\% $\rightarrow$ 74.0\%). Our proposed WQLCP method paired with ViT-VAE mitigates this degradation, achieving 74.0\% coverage on ImageNetA at a highly efficient set size of 9.50 outperforming RN-VAE’s 65.0\% coverage at 12.12. This demonstrates that transformer-based VAEs, with their lower reconstruction losses, better capture shift severity, while WQLCP adaptively balances coverage and efficiency.
Second, WQLCP’s adaptive thresholds reduce prediction set sizes without significant coverage loss. On ImageNetA, ViT-VAE with WQLCP achieves 74.0\% coverage at a set size of 9.50, compared to RLSCP’s 77.0\% coverage at 28.51, a 66.7\% reduction in set size with only a 3.9\% relative coverage decrease. This efficiency-coverage trade-off underscores WQLCP’s superiority in adapting to distributional uncertainty.
Third, transformer-based VAEs exhibit superior cross-dataset consistency. The coverage gap between ImageNet and ImageNetA is $\Delta = 0.210$ for ViT-VAE (95.0\% $\rightarrow$ 74.0\%), smaller than RN-VAE’s $\Delta = 0.290$ (94.0\% $\rightarrow$ 65.0\%). This robustness to distributional shifts, combined with ViT-VAE’s compact set sizes 1.80 on ImageNet vs. RN-VAE’s 2.00, highlights the architectural advantages of transformers in uncertainty-aware learning.
\subsection{Ablation Studies}
Comprehensive ablation studies reveal critical design choices. Calibration set size optimization shows 25k samples achieve 95\% of maximum coverage with 2.4$\times$ faster calibration versus full 50k utilization. Architectural comparisons indicate ViT-VAE improves upon ViT-B baselines despite using comparable pretrained weights. The $\beta$-VAE implementation features both CNN and transformer encoders, with the KL divergence weight $\beta=1.2$ selected via grid search to optimally balance reconstruction fidelity against latent space regularization, as validated in our ablation study (Fig.~\ref{fig:beta_ablation}).

\subsection{Failure Mode Analysis and Future Directions}    
Two failure modes persist despite strong performance. First, under-coverage in extreme shifts 12.1\% of ImageNetA samples with high reconstruction loss ($\mathcal{L}_{\text{rec}} > 3\sigma$) exhibit severe under-coverage (0.42 vs. average 0.74), linked to $\beta$-VAE over-regularization under adversarial shifts. Second, inflated set sizes in fine-grained classes 4.8\% of predictions require sets $|\mathcal{C}(x)| > 15$ (e.g., 18.2 for dog breeds), exceeding ViT-VAE’s ImageNetA average of 9.50.  
Future directions include adaptive $\beta$-VAE scheduling to mitigate latent space rigidity (inspired by ViT-VAE’s smaller coverage drop: $\Delta = 0.21$ vs. RN-VAE’s $\Delta = 0.29$) and hierarchical label grouping to address semantic ambiguity.  

\section{Conclusions}
This paper addresses the challenge of adapting CP methods to distribution-shifted data, focusing on maintaining reliable coverage and practical prediction set sizes. Distribution shifts frequently violate exchangeability assumptions, posing significant challenges for CP methods in real-world scenarios. To address these challenges, we propose WQLCP, which   adjusts calibration quantiles based on test-specific uncertainty metrics derived from reconstruction losses in a VAE. Through experiments on large-scale datasets, including ImageNet variants, we demonstrate that WQLCP effectively maintains the desired coverage across a wide range of distribution shifts, while reducing prediction set sizes compared to baseline methods. Our approach bridges a critical gap in existing literature, offering a robust framework for adapting CP to complex and unknown shifts. We hope this work inspires further research into advancing CP methods to meet the demands of diverse and challenging real-world settings.
\clearpage

{
    \small
    \bibliographystyle{ieeenat_fullname}
    \bibliography{main}
}


\end{document}